%% file: main.tex
\documentclass[runningheads]{llncs}
\usepackage{graphicx}
\usepackage{booktabs}
\usepackage{hhline}
\usepackage{tabularx}
\usepackage{makecell} 
\usepackage{multirow}
\usepackage{comment}

\usepackage{comment}

\usepackage[numbers, sort&compress, sectionbib]{natbib}

\usepackage{amsmath}

\usepackage[misc,geometry]{ifsym} 

\DeclareMathOperator{\softmax}{softmax}
\DeclareMathOperator*{\logits}{logits}

\usepackage[table]{xcolor}

\def\HS{\hspace{\fontdimen2\font}}

\makeatletter
\def\thebibliography#1{\section*{References\@mkboth
    {REFERENCES}{REFERENCES}}\list
    {[\arabic{enumi}]}{\settowidth\labelwidth{[#1]}\leftmargin\labelwidth
\advance\leftmargin\labelsep
\usecounter{enumi}}
\def\newblock{\hskip .11em plus .33em minus .07em}
\sloppy\clubpenalty4000\widowpenalty4000
\sfcode`\.=1000\relax}
\makeatother

\begin{document}
\title{Machine learning for dynamically predicting the onset of renal replacement therapy in chronic kidney disease patients using claims data}
\titlerunning{ML for predicting RRT onset in CKD patients using claims data}

\author{Daniel Lopez-Martinez$^{(\textrm{\Letter})}$ \and
Christina Chen \and
Ming-Jun Chen}


\authorrunning{D. Lopez-Martinez et al.}

\institute{Google Research, Palo Alto, USA \\
\email{dlmocdm@google.com} }

\maketitle              

\begin{abstract}
Chronic kidney disease (CKD) represents a slowly progressive disorder that can eventually require renal replacement therapy (RRT) including dialysis or renal transplantation. Early identification of patients who will require RRT (as much as 1 year in advance) improves patient outcomes, for example by allowing higher-quality vascular access for dialysis. Therefore, early recognition of the need for RRT by care teams is key to successfully managing the disease. Unfortunately, there is currently no commonly used predictive tool for RRT initiation. In this work, we present a machine learning model that dynamically identifies CKD patients at risk of requiring RRT up to one year in advance using only claims data. To evaluate the model, we studied approximately 3 million Medicare beneficiaries for which we made over 8 million predictions. We showed that the model can identify at risk patients with over 90\% sensitivity and specificity. Although additional work is required before this approach is ready for clinical use, this study provides a basis for a screening tool to identify patients at risk within a time window that enables early proactive interventions intended to improve RRT outcomes.
\keywords{Dialysis  \and Kidney transplant \and Chronic kidney disease \and Claims data \and Dynamic modeling}
\end{abstract}

\input{sections/introduction}
\input{sections/methods}
\input{sections/results}
\input{sections/discussion}

\bibliographystyle{splncs04}
\bibliography{paperpile}

\end{document}

%% file: sections/introduction.tex
\section{Introduction}
Chronic kidney disease (CKD) is defined by decreased kidney function for three or more months and is classified based on the severity of kidney damage. As CKD is not reversible, the natural course is progression over time with the most severe form being end-stage renal disease (ESRD). When disease progresses to ESRD, residual kidney function is no longer able to meet the body’s needs without the aid of life-sustaining chronic renal placement therapy (RRT), which includes chronic dialysis (hemodialysis or peritoneal dialysis) and kidney transplant.  CKD affects approximately one-seventh of US adults above the age of 20 years and constitutes one of the major public health problems globally \cite{Coresh2007-rg}. In the United States, there are 37 million patients with CKD and over 700,000 who have progressed to ESRD \cite{Coresh2007-rg, noauthor_undated-gj}.   

The development of ESRD is associated with significant morbidity and mortality, and the Kidney Disease Outcomes Quality Initiative (KDIGO)  \cite{Wavamunno2005-qi} has emphasized the importance of an individualized approach to patient care, taking into account life expectancy, comorbidities, individual vascular characteristics, as well as individual patient circumstances, needs, and preferences. Ideally, these multidisciplinary discussions should start at least 9-12 months prior to RRT initiation to allow time for adequate discussion and if appropriate, dialysis access (arteriovenous fistula or graft or peritoneal dialysis catheter) creation and maturation, and kidney transplant evaluation and listing. Dialysis access requires surgical intervention and takes time before it is ready for use; patients who do not receive adequate preparation require placement of central venous catheters for dialysis, and may be exposed to complications like severe infections \cite{Kusminsky2007-nw}. 

Identifying patients at risk of developing ESRD and requiring RRT would enable care teams to target these populations to initiate multidisciplinary discussions, which have been associated with improved survival, lower hospitalization rates, improved uptake of dialysis and better access to kidney transplant waiting lists  \cite{Smart2011-il, Smart2014-wj, Hurst2020-oq, Koyner2018-yg}. Unfortunately, it is difficult to predict the development of ESRD and most patients do not receive the recommended clinical care before RRT initiation \cite{Huang2014-dr, Mehrotra2005-ab}. Despite substantial efforts (e.g. Fistula First Catheter Last \cite{Lee2017-hf}), there has been virtually no reduction in central venous catheter use at initiation of hemodialysis over the last decade \cite{Lee2017-hf, Johansen2021-zs}. Only 20\% of patients  start dialysis through the recommended type of access, an arteriovenous fistula \cite{Lee2017-hf}. 

While there have been efforts to build predictive algorithms to identify at risk patients, there remains no standardized nationally implementable prediction models. Some prior studies that have modeled CKD progression include linear models based on static variables at single points in time, such as the Tangri et al model \cite{Tangri2011-xr}, which have been evaluated in multiple studies \cite{Tangri2011-xr, Tangri2016-ay}. While promising and well studied, these static models do not take into account the temporal progression of disease. Only a limited number of studies have focused on predicting disease deterioration with prediction horizons in the order of months \cite{Norouzi2016-mg}. While these studies have yielded promising results, they relied on medical record data, including laboratory values, which are not as widely available as claims data. To this end, Dovgan et al. \cite{Dovgan2020-uj} developed a model based solely on the comorbidities data from the National Health Insurance in Taiwan to predict initiation of RRT 3, 6, and 12 months from the time of the patient’s first diagnosis with CKD.  While this study showed promising results, predictions were only rendered once at incident diagnosis of CKD without any additional evaluation of subsequent disease progression. 

To address the aforementioned limitations, in this study we describe a time-bucketed linear machine learning model to predict RRT initiation in patients with a diagnosis of CKD in a format that is structured to be deployable at a large scale: 1) Use of nationally available Medicare claims; 2) Integrating temporal information; 3)  A linear model that is easy to interpret; 4) Monthly predictions to match the realistic cadence of care management groups. We evaluate the performance of our model on prediction windows of up to 365 days. In addition to this, we also evaluate the potential impact our model may have on improving timely dialysis access prior to initiation of RRT.

%% file: sections/methods.tex
\section{Methods}

\subsection{Dataset description} \label{sec:dataset_description}
We used the Limited Dataset (LDS) Standard Analytic Files (SAFs) sample of Medicare beneficiaries from the Centers for Medicare \& Medicaid Services (CMS), the largest insurer of people aged 65 years and older in the USA.  This dataset contains a 5\% random sample of beneficiaries enrolled in the fee-for-service Medicare program that is representative of the overall Medicare population, and their corresponding Part A (inpatient/hospital coverage) and Part B (outpatient/medical coverage) fee-for-service administrative claims from January 1, 2011 to December 31, 2016.
We included all patients who at any point after turning 65 had a fee for service claim, regardless of the original enrollment reason, whether they switched from a non fee for service plan, or they died. 

\subsection{Task definition} \label{sec:task_definition}
Our goal was to identify patients at risk of requiring renal replacement therapy, specifically (1) chronic dialysis, (2) kidney transplant, and (3) any renal replacement therapy (RRT) defined as either of the previous two. We specifically sought to predict the onset of these three therapy categories within the following clinically relevant time horizons: 30, 60, 90, 180, 365 days. 

The definitions of chronic dialysis, renal transplant, and RRT are derived from Healthcare Common Procedure Coding Systems (HCPCS) procedure codes,
which are included in the claims data. A first occurrence in a beneficiary’s claims record of any of the corresponding procedure codes would denote initiation of the corresponding therapy.

\subsection{Data representation and processing} \label{sec:data_representation}

\subsubsection{Data Assembling}
The dataset described in Sec.\ref{sec:dataset_description} contains claims of the following types: inpatient, outpatient, home health, skilled nursing, and non-institutional (i.e. carrier). For each beneficiary in the dataset, we assembled these to produce a temporally ordered list of claims. This list was sampled periodically to generate machine learning training samples and render predictions. We call this process “triggering”. In order to simulate care management workflows, eligible predictions for all beneficiaries were triggered on the first of every month between January 1, 2012 and December 1, 2015.  This provides one year of buffer after the last potential trigger, as the dataset ended in December  31,  2016. 

For each trigger, we assembled the entire sequence of claims up to the first of the month. Each trigger time was associated with a binary label for each of the three tasks (dialysis, kidney transplant, renal replacement) and the five time horizons (30, 60, 90, 180, 365 days). If any of the procedure codes corresponding to the task (see Sec.\ref{sec:task_definition}) was found in the beneficiary’s claims within the corresponding time horizon from trigger time, a positive label was generated. Otherwise, it was negative.

\subsubsection{Prediction Triggering Eligibility Criteria}
\begin{figure}[hbt!]
\includegraphics[width=1.0\linewidth]{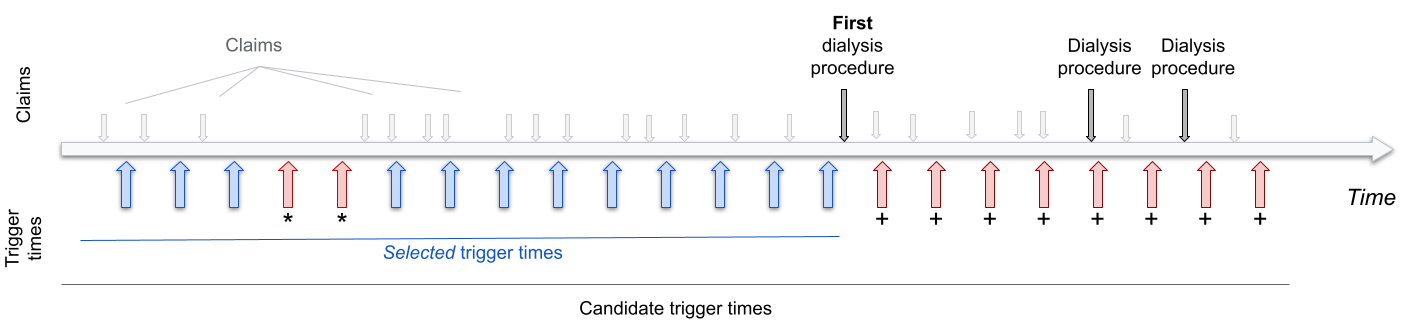}
\caption{Example triggers events for a patient. Triggering occurs on the first of every month as represented by the lower blue and red arrows. Claim events, including claims correlated with chronic dialysis procedures, are represented by the gray upper arrows. If all inclusion criteria are met for a patient at the time of triggering, a prediction of dialysis risk within the horizon window (next 30, 60, 90, 180, or 365 days) will be generated (blue arrow: prediction generated, red arrow: prediction not generated). For example, predictions are not generated if no claims have been filed in the previous 30 days (red arrows marked with *). Triggers after the first chronic dialysis procedure will not generate a prediction (red arrows marked with a +).}
\label{fig:triggering}
\end{figure}

Because of how the machine learning problem was formulated, not all possible triggers were used by the model. Candidate triggers were eligible for predictions if at the time of triggering all the following criteria were met: the patient (1) was 65 years or greater; (2) had diagnosis of CKD on any previous claim; (3) had not initiated RRT; (4) had at least  a year of historical claims; (5) had a healthcare claim within the last 30 days (to avoid rendering new predictions for patients whose claims data does not contain any new information since the last prediction). Only triggers that met the aforementioned conditions were used by the model for either training or testing. This trigger filtering process is depicted in Fig.\ref{fig:triggering}.

\subsubsection{Feature Extraction}

We implemented a data representation that has been shown to accurately predict multiple medical events while requiring minimal data processing for improved generability and ease of implementation \cite{Rajkomar2018-kv}. 

For each beneficiary, we extracted demographic data from the initial enrollment into Medicare. Specifically, recorded sex, reported race, and birth-year which was then converted to bucketed age (with 10-year buckets) at each trigger time. In addition to this, for each calendar day, we extracted information about the conditions, procedures, encounter types, and administered medications from the inpatient, outpatient, home health, skilled nursing, and non-institutional (i.e. carrier) claims. This information was used to develop a total of 23 features with one-hot encoding, indicated in Table \ref{tab:features}.

\begin{table}[hbt!] \centering
\begin{tabular}{|l|p{8cm}|}
\hline
\textbf{Feature group (\# feat.)} & \textbf{Features} \\ \hline
Patient characteristics (3) &       sex, race, bucketed age (10-year buckets) \\ \hline
Conditions (4) &                    ICD9, ICD10, CCS, HCC  \\ \hline
Procedures (6) &                    ICD9, ICD10, CCS, CPT, hcpcs.alpha, performer role \\ \hline
Encounters (8) &                    class, hospitalization admit source, discharge disposition, ICD9, ICD10, CCS, HCC, code revenue type \\ \hline
Medications (2) &                   HCPCS, rxnorm \\ \hline
\end{tabular}
\caption{List of the 23 features used in the model grouped by type, together with the number of features in each group.}
\label{tab:features}
\end{table}

The condition-based features used all the raw ICD-9 and ICD-10 codes associated with each claim as well as the Clinical Classification Software (CCS) and Risk Adjustment and Hierarchical Condition Category (HCC) codes. The procedure-based features included the HCPCS codes from the claims, as well as ICD-9, ICD-10, CCS, and Current Procedural Terminology (CPT) codes, which were derived from the HCPCS codes, in addition to the performer role (indicating that a procedure was done by a provider in e.g. "General Surgery" or "Urology"). The encounter-based features indicated the class of encounter (e.g. inpatient, ambulatory, etc), the encounter source code for inpatient admissions, the discharge disposition of the encounter, the ICD-9 or ICD-10 principal diagnosis in the encounter, and the revenue center codes. Finally, the medication administration features included HCPCS codes as well as RXNorm codes, which were derived from the National Drug Codes (NDCs).

Note that during the time period covered in this dataset, there was a gradual transition from ICD-9 to the updated ICD-10 billing codes and therefore both coding schemes were included.

With the exception of the demographic data, all feature values that occurred within a time range were aggregated into a new feature. We refer to this process as time-bucketing. In this work, we used the following non-overlapping disjoint time buckets: [0, 30 days), [30 days, 90 days), [90 days, 1 year), [1 year, 10 years).

\subsubsection{Dataset split}
To generate train, validation, and test datasets for modeling we split the entire dataset by beneficiary using a random split of 80\%/10\%/10\% respectively.  There is no overlap of beneficiaries in these three groups.

\subsection{Model description}

Using the features described in 4.2, we designed independent logistic regression machine learning models to render predictions for each of the three tasks (chronic dialysis, kidney transplant, and RRT) for each beneficiary on the 1st day of every month (if eligible based on the triggering eligibility criteria; see Sec.\ref{sec:data_representation}) using the data available up to that point.  

At training time, we cast the problem as multiclass by dividing our overlapping prediction windows (0-30, 0-60, 0-90, 0-180, 0-365 days from trigger time) into disjoint windows (0-30, 30-60, 60-90, 90-180, 180-365 days from trigger time). In this setting, only the disjoint window that contained the event is positive, while all others are negative. Additionally, we append one extra output to explicitly represent the negative class (i.e. examples in which no adverse event occurred in any prediction window). For example, with 30/60/90/180/365-day windows, the label for an example with an event at day 50 would be [0,1,0,0,0,0], whereas the label for an example with no events within 365 days of prediction would be [0,0,0,0,0,1]. In this setting, we compute scores as softmax(logits) and train the model using a softmax cross-entropy loss. However, at inference time, we recover predictions for the overlapping windows by computing cumulative sum probabilities: for window $i$, we compute its probability as $p_i=\sum_{j=0}^{i} s_j$, where $s_j=\softmax(\logits[j])$. This constraint ensures that the output probabilities monotonically increase over successive prediction windows.

All models were implemented and trained with TensorFlow in Python \cite{Abadi2016-it}. Model hyperparameters, including L1 regularization coefficient and the parameters of an exponential learning rate decay, were tuned by minimizing loss on the validation set.

\subsection{Model Evaluation}
We measured the performance of the model on an independent held out test set comprising 10\% of the available beneficiaries. No beneficiaries in the test set were used for model training or validation. For each prediction horizon, we calculated the receiver operating characteristic area under the curve (ROC-AUC) and the precision-recall AUC (PR-AUC). Note that given the low prevalence of triggers with positive labels, the PR-AUC is more informative \cite{Saito2015-oi}. In addition to this, we also computed the sensitivity and specificity by selecting the threshold that maximized the geometric mean.


Finally, we evaluated the potential impact of our model in improving rates of dialysis access creation (arteriovenous fistula creation, arteriovenous graft placement, and peritoneal dialysis catheter placement). Dialysis access creation was defined by the corresponding list of procedure codes. Specifically, we determined the number of patients we identified using a 1 year prediction horizon who had not undergone dialysis access creation before dialysis initiation, at three target sensitivities: 60\%, 70\%, 80\%.

%% file: sections/results.tex
\section{Experiments and Results}

\subsection{Study population and dataset}

A total of 2,978,542 individuals were present in the dataset. Their gender and race statistics are shown in Table 1. Of these, 572,319 (19.22\%) received a diagnosis with CKD at some point as indicated by the corresponding ICD9 and ICD10 codes in their claims history. We used these codes to build a cohort of CKD patients in which to render predictions for the onset of RRT.  The statistics for the CMS dataset and the cohort of CKD patients is shown in Table \ref{tab:table1}.

\begin{table}[] \centering 
\begin{tabular}{r|rl|rl|}
\cline{2-5}
\multicolumn{1}{l|}{}                           & \multicolumn{2}{c|}{\HS\HS\HS Development set \HS\HS\HS} & \multicolumn{2}{c|}{\HS\HS\HS\HS\HS\HS\HS Test set \HS\HS\HS\HS\HS\HS\HS }  \\ \hline \hline
\multicolumn{1}{|l|}{\textbf{Patients}}         & 325,851            &                 & 36,077               &         \\ \hline \hline
\multicolumn{1}{|l|}{Female, n (\%)}            & 174,851            & (53.66\%)         & 19,357               & (53.65\%) \\ \hline
\multicolumn{1}{|l|}{Race, n (\%)}              &                    &                 &                      &         \\ \hline
\multicolumn{1}{|l|}{Asian}                     & 7,007              & (2.15\%)          & 786                  & (2.18\%)  \\ \hline
\multicolumn{1}{|r|}{\textit{Black}}            & 36,155             & (11.10\%)         & 4,042                & (11.20\%) \\ \hline
\multicolumn{1}{|r|}{\textit{Hispanic}}         & 6,572              & (2.02\%)          & 789                  & (2.19\%)  \\ \hline
\multicolumn{1}{|r|}{\textit{Native American}}  & 1,508              & (0.46\%)          & 161                  & (0.45\%)  \\ \hline
\multicolumn{1}{|r|}{\textit{White}}            & 268,652            & (82.45\%)         & 29,656               & (82.20\%) \\ \hline
\multicolumn{1}{|r|}{\textit{Other}}            & 4,600              & (1.41\%)          & 486                  & (1.35\%)  \\ \hline
\multicolumn{1}{|r|}{\textit{Unknown}}          & 1,357              & (0.42\%)          & 157                  & (0.44\%)  \\ \hline \hline
\multicolumn{1}{|l|}{\textbf{Predictions}}      & 7,772,847          &                 & 704,983              &         \\ \hline \hline
\multicolumn{1}{|l|}{Female, n (\%)}            & 4,244,917          & (54.61\%)         & 387,741              & (55.00\%) \\ \hline
\multicolumn{1}{|l|}{Race, n (\%)}              &                    &                 & \multicolumn{1}{l}{} &         \\ \hline
\multicolumn{1}{|r|}{\textit{Asian}}            & 155,521            & (2.00\%)          & 14,414               & (2.04\%)  \\ \hline
\multicolumn{1}{|r|}{\textit{Black}}            & 826,666            & (10.64\%)         & 74,413               & (10.56\%) \\ \hline
\multicolumn{1}{|r|}{\textit{Hispanic}}         & 147,134            & (1.89\%)          & 14,419               & (2.05\%)  \\ \hline
\multicolumn{1}{|r|}{\textit{Native American}}  & 33,532             & (0.43\%)          & 3,058                & (0.43\%)  \\ \hline
\multicolumn{1}{|r|}{\textit{White}}            & 6,484,990          & (83.43\%)         & 587,145              & (83.28\%) \\ \hline
\multicolumn{1}{|r|}{\textit{Other}}            & 102,687            & (1.32\%)          & 9,114                & (1.29\%)  \\ \hline
\multicolumn{1}{|r|}{\textit{Unknown}}          & 22,317             & (0.29\%)          & 2,420                & (0.34\%  \\ \hline
\multicolumn{1}{|l|}{Age, n (\%)}               &                    &                 &                      &         \\ \hline
\multicolumn{1}{|r|}{\textit{65-74 years}}      & 2,019,746          & (31.83\%)         & 225,559              & (31.99\%) \\ \hline
\multicolumn{1}{|r|}{\textit{75-84 years}}      & 2,602,663          & (41.01\%)         & 286,072              & (40.58\%) \\ \hline
\multicolumn{1}{|l|}{\textit{\HS\HS85 years or more}} & 1,723,650          & (27.16\%)         & 193,352              & (27.43\%) \\ \hline
\end{tabular}
\caption{Statistics of the development (train and validation) and test datasets. Note that the train, validation and test split were done according to beneficiary id on the entire dataset to accomplish an 80/10/10 distribution. Here, we include the distribution of patients after triggering has been performed, which may exclude some patients if none of their triggers meet the trigger eligibility criteria.}
\label{tab:table1}
\end{table}

\subsection{Model performance}

Table \ref{tab:performance_ckd_cohort} compares the performance of all tasks for 5 overlapping windows: 30, 60, 90, 180, 365 days from trigger time. The best performance was achieved for the renal replacement task, which was similar to the chronic dialysis task as most patients in the dataset are  dialysis patients. However, merging the chronic dialysis and kidney transplant tasks did improve the model performance slightly. The kidney transplant task had the lowest model performance, which is explained by only 0.71\% patients on the CKD cohort having a kidney transplant (versus 4.88\%) and the overall low label prevalence for all predictions across all horizons (see Table \ref{tab:label_prevalence}).
Performance of all metrics decreased with increasing prediction horizons, reflecting the increased difficulty in accurately predicting onset of therapy for large prediction horizons.
Note that given the low prevalence of triggers with positive labels, the PR-AUC is more informative \cite{Saito2015-oi}.

\begin{table}[t] \centering
\begin{tabular}{|l|r|r|r|}
\hline
Prediction Horizon & \multicolumn{1}{c|}{\HS\HS\HS\HS\HS\HS\HS\HS   RRT \HS\HS\HS\HS\HS\HS\HS\HS  } & \multicolumn{1}{c|}{\HS\HS Chronic Dialysis \HS\HS } & \multicolumn{1}{c|}{ \HS\HS Kidney Transplant \HS\HS } \\ \hline \hline
30d                & 0.09\%                   & 0.09\%                                & 0.01\%                                 \\ \hline
60d                & 0.18\%                   & 0.18\%                                & 0.01\%                                 \\ \hline
90d                & 0.27\%                   & 0.26\%                                & 0.02\%                                 \\ \hline
180d               & 0.53\%                   & 0.52\%                                & 0.04\%                                 \\ \hline
365d               & 1.06\%                   & 1.05\%                                & 0.07\%                                 \\ \hline
\end{tabular}
\caption{Label prevalence for the three prediction tasks and each prediction horizon.}
\label{tab:label_prevalence}

\vspace{0.2cm} %

\begin{tabular}{|l|l|l|l|l|l|l|}
\hline
Task                                         & Metric      & next 30d & next 60d & next 90d & next 180d & next 365d \\ \hline \hline
\multirow{4}{*}{\parbox{1.8cm}{Renal \\replacement\\procedure}} & ROC-AUC     & 0.971        & 0.963        & 0.953        & 0.942         & 0.928         \\ \cline{2-7} 
                                             & PR-AUC      & 0.311        & 0.285        & 0.261        & 0.262         & 0.280         \\ \cline{2-7} 
                                             & Sensitivity & 0.927        & 0.906        & 0.885        & 0.870         & 0.865         \\ \cline{2-7} 
                                             & Specificity & 0.940        & 0.922        & 0.910        & 0.887         & 0.855         \\ \hline \hline
\multirow{4}{*}{\parbox{1.8cm}{Dialysis\\procedure}}          & ROC-AUC     & 0.974        & 0.964        & 0.951        & 0.941         & 0.927         \\ \cline{2-7} 
                                             & PR-AUC      & 0.172        & 0.148        & 0.131        & 0.119         & 0.117         \\ \cline{2-7} 
                                             & Sensitivity & 0.904        & 0.921        & 0.882        & 0.862         & 0.845         \\ \cline{2-7} 
                                             & Specificity & 0.951        & 0.895        & 0.894        & 0.880         & 0.858         \\ \hline \hline
\multirow{4}{*}{\parbox{1.8cm}{Kidney\\transplant\\procedure}} & ROC-AUC     & 0.975        & 0.975        & 0.975        & 0.972         & 0.961         \\ \cline{2-7} 
                                             & PR-AUC      & 0.006        & 0.011        & 0.014        & 0.032         & 0.047         \\ \cline{2-7} 
                                             & Sensitivity & 0.930        & 0.919        & 0.937        & 0.923         & 0.903         \\ \cline{2-7} 
                                             & Specificity & 0.939        & 0.941        & 0.922        & 0.923         & 0.918         \\ \hline
\end{tabular}
\caption{Comparison of ROC-AUC, PR-AUC, sensitivity, and specificity for the three prediction tasks in the cohort of CKD patients. Sensitivity and specificity was computed by choosing the threshold that maximized the geometric mean.}
\label{tab:performance_ckd_cohort}

\vspace{0.2cm} %

\begin{tabular}{|r|r|r|}
\hline
\HS Sensitivity \HS & \HS Specificity \HS & \HS \% of patients \HS \\ \hline
80\%        & 91.62\%     & 35.36\%        \\
70\%        & 95.18\%     & 33.49\%        \\
60\%        & 97.32\%     & 31.80\%       \\ \hline
\end{tabular}
\caption{Percentage of beneficiaries identified by our model using a 1 year prediction horizon that had not undergone dialysis access creation procedures before dialysis initiation, for three target sensitivities and their corresponding specificities.}
\label{tab:performance_impact}
\end{table}

In addition to this, Table \ref{tab:performance_impact} shows the percentage of beneficiaries that had been identified correctly to start dialysis within the 1 year prediction horizon that had not undergone  dialysis  access  creation  procedures  before  dialysis initiation, hence illustrating the potential impact of the model.

%% file: sections/discussion.tex
\section{Conclusions}

We developed a machine learning model for dynamically identifying those patients with high likelihood of starting RRT up to a year in advance. Our study has limitations. First, it was conducted with a subset of CMS patients and therefore this approach should be tested in other health systems. Second, the model used claims data only. While this may help with adoption as this information is generally available, access to laboratory data (e.g. glomerular filtration rate or albuminuria), patient’s characteristics such as weight or the complete electronic health record could significantly improve the model. Thirdly, we recognize that models based on claims are intrinsically reliant on healthcare billing patterns which may be associated with racial biases due to unequal historical access. Until recently, definitions of CKD were based on race-based formulae; the new guidelines recommend a race agnostic formula \cite{Inker2021-yr, Delgado2021-ap}. Further work is needed to characterize and mitigate potential biases in the CKD use-case.  Finally, this was a retrospective study, and therefore the model should be evaluated in a prospective study with a heterogeneous group of patients before adoption in clinical practice.

In addition to this, note that  model performance may be improved by implementing more complex modeling architectures (e.g. neural networks), addressing class imbalance by upsampling  the unrepresented classes or using weighted loss functions, and reducing the dimensionality of the feature space by including only the most relevant features. These can be identified through feature attribution methods, which can also be used to improve the interpretability of complex models such as neural networks.

\textbf{Prospect of application:} It is difficult for providers to predict when patients with chronic kidney disease will need renal replacement therapy. Emergency renal replacement therapy initiations increase morbidity and mortality and cause significant stress. Our AI model identifies patients at risk of progression and can help care management groups systemically focus on these patients to physically and emotionally prepare them for dialysis initiation.